# AI Enhanced Multi-Robotic Systems for inpatient care and Diagnostic support

Nakhul Kalaivanan
Robotics, Engineering and Physical Sciences
, *Heriot-Watt University*
*Dubai, United Arab Emirates*
nk2002@hw.ac.uk

Girish Balasubramanian,
Engineering and Physical Sciences
, *Heriot-Watt University*
*Dubai, United Arab Emirates*
gb2018@hw.ac.uk

Senthil Muthukumaraswamy,
Engineering and Physical Sciences
, *Heriot-Watt University*
*Dubai, United Arab Emirates*
sam58@hw.ac.uk

*Abstract*—This research presents a multi-robot system for inpatient care, designed using swarm intelligence principles and incorporating wearable health sensors, RF-based communication, and AI-driven decision support. Within a simulated hospital environment, the system adopts a leader–follower swarm configuration to perform patient monitoring, medicine delivery, and emergency assistance. Due to ethical constraints, live patient trials were not conducted; instead, validation was carried out through controlled self-testing with wearable sensors. The Leader Robot acquires key physiological parameters, including temperature, $SpO_2$, heart rate, and fall detection, and coordinates other robots when required. The Assistant Robot patrols corridors for medicine delivery, while a robotic arm provides direct drug administration. The swarm-inspired leader–follower strategy enhanced communication reliability and ensured continuous monitoring, including automated email alerts to healthcare staff. The system hardware was implemented using Arduino, Raspberry Pi, NRF24L01 RF modules, and a HuskyLens AI camera. Experimental evaluation showed an overall sensor accuracy above 94%, a 92% task-level success rate, and a 96% communication reliability rate, demonstrating system robustness. Furthermore, the AI-enabled decision support was able to provide early warnings of abnormal health conditions, highlighting the potential of the system as a cost-effective solution for hospital automation and patient safety.

I. INTRODUCTION

Hospitals today face a range of systemic challenges, including chronic staff shortages, rising patient numbers, and the increasing demand for continuous monitoring of vital signs. These pressures place a considerable burden on healthcare providers and often compromise the speed and quality of patient care.. Such challenges cause a delay in getting the medicines, slower assistance in emergencies, and more stress on the health workers. Robotic systems may help with dedicated duties and complement medical personnel, which allows faster and more reliable patient care. Social insects often inspire swarm robotics and can enable many small robots to work together to perform complex tasks and complete objectives. Swarm robots can help throughout the healthcare sector, ranging from drug delivery to hospital cleaning and patient monitoring. Swarm robots are decentralised systems that can adapt to dynamic environments and scale efficiently. The motivation for this work is deeply personal. This project is for my grandmother, who passed away, because help did not arrive quickly enough for a proper diagnosis and treatment. This incident highlighted the need for a system that allows for continuous patient monitoring and the prompt administration of medication. An AI-enhanced multi-robot system for inpatient care is presented in this paper. The system uses a leader-follower approach whereby the Leader Robot collects patient data from wearing devices and commands the other robots.

The Assistant Robots walk down the halls and deliver medicines, while the robotic arm dispenses drugs immediately. AI gets inserted as a decision-support layer that analyses patient vitals and provides in-time warnings of health problems. The purpose of this work is to develop a dependable and expandable robot framework. It will enhance patient monitoring, reduce response time in emergencies and assist the hospital staff in regular operations. In a controlled environment, testing was done using wearable sensors, RF modules, Arduino, Raspberry Pi, and HuskyLens AI cameras.[1-7]In the study, effective communication, accurate monitoring, and successful drug delivery were obtained, which proves that we could achieve this performance in real life in the future. Main Objectives of this Paper:

- Design and deploy a Leader Follower swarm robotic system using RF communication and modular robots for inpatient care.

- Integrate wearable device or sensor data like BPM, temperature, and oxygen saturation for real-time patient condition monitoring.

- Enabling AI to analyse patient vitals and assist them whether they are at home or in the hospital. The diagnosis and treatment plans.

- Measure the system effectiveness in terms of response time, ethical safety, communication stability, and adaptability under emergency or non-emergency conditions. This includes ensuring timely medical delivery as scheduled by doctors via the Blynk app, with the Corridor Robot monitoring for falls during patrols, while maintaining patient safety, stable communication, and faster coordination between all robotic units.
- AI-Based Algorithms for offline machine learning analysis.

II. HARDWARE COMPONENTS

The system was built using simple and low-cost hardware parts that can work together for patient monitoring, medicine delivery, and communication. The Components used in the system are mentioned below[8]:

A. *Arduino Mega 2560 and Arduino Uno*

The Arduino Mega 2560 and Uno boards serve as the main controllers of the robots. These microcontrollers are programmed in C/C++ and are responsible for handling sensor data, motor control, and wireless communication. The Arduino Mega, with 54 digital and 16 analogue pins, was used for tasks requiring multiple connections, while the Uno was used for lighter modules[10].

B. *NRF24L01 Wireless Module*

The NRF24L01 module links the Leader Robot, the Assistant Robot, and the Robotic Arm using RF communication. The low-cost transceiver ensures reliable and fast data transmission with low power consumption for continuous monitoring in the hospital[11].

C. *Wearable sensors*

The Wearable device contains a pulse oximeter ($SpO_2$), heart rate sensor (BPM), temperature sensor, and falling sensor. The modules wirelessly transmit real-time vital data from the patient to the Leader Robot for analysis[11].

D. *Raspberry PI 5 With ChatGPT Integration*

The Raspberry Pi 5 is the processor for advanced decision-making based on AI. It is configured to process patient vitals, perform analysis in real-time and provide diagnostic support. Using API keys, the Raspberry Pi 5 has been integrated with a ChatGPT-based AI model to read health data, provide recommendations, and communicate with patients through a chatbot interface on a 7-inch screen. Also, the system sends automated alerts and emails to doctors. This helps in a quick response to emergencies. It also improves communication between patients and medical staff[12].

E. *Husky Lens AI camera*

The Husky lens AI camera is used for vision-related tasks like paitent identification, corridor patrolling, and fall detection. It performs many AI functions, like recognising objects, detecting faces, and detecting lines, which is suitable for monitoring patient movement in the hospital. Assistant Robot uses HuskyLens to scan the corridor, detect a fallen patient and deliver medicine. Through visual monitoring and decision support, this module decreases dependency on manual supervision. Thus, it enables the swarm robotic system[12].

F. *DC Motor(JGB37-520 ) and Calibration*

The JGB37-520 DC motor helps the Leader and Assistant robots of the hospital to move smoothly. It is a high-torque, low-speed motor well suited for line-following and mobile robots. The H-bridge driver controls the motor to move forward and reverse. Calibration is done by changing the motor speed using Pulse Width Modulation (PWM) signals, and comparing the movement of the robot against the required path. To correct the speed and turning of the robot during navigation, the sensor relies on encoders and potentiometers. Reduces the rate of error during navigation.

G. *5 Way IR sensor*

The infrared (IR) sensor array of the robot work together in conjunction and are mounted on the front underside of the chassis to allow line-following navigation. The module has five sets of infrared transmitter-receivers in a straight line. Every sensor measures the floor's surface. By emitting infrared light, it reflects[15-19].

Black line → absorbs IR light → produces a low output.

White surface → reflects IR light → produces a high output.

The robot can detect the position of the line with respect to the robot chassis. The labels on the left and right sensors are Left-2,Left-1 Centre,Right-1 , and Right-2. The drifter robot can drift to the left or right side of the path.

In hardware design, the digital output pins of the sensor module will be connected to the Arduino Mega 2560. The Arduino gets binary signals (either 1 or 0) from each of the sensors. It will process them and provide the required corrections through the motor driver for steering. The robot's speed will be regulated along the path by the DC motors (JGB37-520).By using the five-way configuration, the robot is stabilised more than with a single IR sensor or a dual IR sensor. The robot is able to sharply turn and take deviations with the help of the hospital floor detection sensor.

H. *Leader Robot System*

The Leader Robot manages and coordinates the system to provide inpatient care. The system gets data from wearable health monitoring sensors such as the $SpO_2$, pulse rate, temperature, and fall detections and processes them in real-time. We utilise a Raspberry Pi 5 for the hardware used in our project. Further, it is powering an AI-based decision-making system. In addition, the system is also integrated with ChatGPT using some API keys. The Leader Robot may use a diagnostic screen, display vital signs on a screen, and communicate with the patient through a chatbot. To coordinate the task, the Leader Robot uses an NRF24L01 RF communication module to command the other units. When an abnormal condition is observed (i.e. low $SpO_2$ or a detected fall), the Leader Robot sends a command to the Corridor Robot (i.e. for emergency patrol, medicine delivery). If medicine is ordered or scheduled, the Robotic Arm Unit picks up and feeds the medicine to the patient. This leader–follower swarm strategy allows monitoring and task function while limiting communication risk[15-19].

I. *Corridor Robot System*

The Corridor Robot patrols the corridors of the simulated hospital, which performs emergency intervention and medicine delivery. It uses four wheels that run on DC motors (JGB37-520) for smooth operation. The motors were calibrated so that the robot could follow lines properly along the hospital track and move between patient rooms. There is a HuskyLens AI vision sensor embedded, and it can be used to detect if a patient is standing or has fallen. With this, the robot is able to recognise emergencies if a user is not wearing the sensor. The robot has an LED tower light, which allows users to easily see it when it is patrolling, delivering medicine, or in emergency mode. The leader robot communicates with the robots through the NRF24L01 module, which provides a fast, low-latency RF transmission. The Leader Robot hands over the assignment to the Corridor Robot for the identification of abnormal conditions from patient data. When given a command, the Corridor Robot will patrol the corridor, check the situation with the HuskyLens and deliver drugs if necessary. The Leader Robot's workload is reduced, and task distribution is assured with this coordination[13].

*J. Robotic Arm with NRF sensors and Blynk Scheduling*

A robotic arm handles medication and takes it to patients, physically. It contains NRF24L01-2 wireless modules, which receive commands from the Leader Robot in the Swarm system. The Blynk IoT platform manages the scheduling of medicine delivery, allowing doctors to set the timing and dosage remotely. After a schedule is triggered, the robotic arm picks up the required medication automatically and delivers it to the patient's bed. The use of NRF sensors for communication and a Blynk-based scheduler minimises the chance of error due to human dependency and also takes care of medicine[14].

III. METHODOLOGY

This section describes the design, development, and testing of the proposed healthcare system utilizing robotic units. The methodology covers the research approach, system workflow, integration of wearable sensors, RF-based communication, AI-driven decision-making, locomotion control, and testing protocols. System validation was performed using quantitative metrics, including response time, sensor accuracy, and task success rate, as well as qualitative observations, such as robot coordination, usability, and communication reliability. These results are presented and analyzed in the subsequent section.[9].

*A. Research Approach*

The project will adopt a pragmatic mixed-methods approach involving hardware prototyping and software integration. Live testing on patients could not be undertaken due to ethical reasons; hence, controlled self-testing using wearable sensors was performed. Simulations included various medical conditions, including abnormal heart rate, falls, and fever. The system is evaluated on quantitative (accuracy, task completion rate, response time) and qualitative (robot coordination, usability, communication reliability) terms[9-19].

*B. System Workflow*

The system is based on a leader–follower swarm architecture. The Leader Robot collects patient vital signs, including $SpO_2$, heart rate (BPM), body temperature, and fall detection data. These measurements are processed by a Raspberry Pi 5 running ChatGPT, which assists in decision-making. Using NRF24L01 RF modules, the Leader Robot assigns tasks to the Corridor Robot and the Robotic Arm. The Corridor Robot navigates hospital corridors and employs a HuskyLens AI camera to detect both standing and fallen patients. It can also deliver or collect medications as required. The Robotic Arm administers drugs according to a predefined schedule or in response to emergency requirements, utilising Blynk IoT for task scheduling and control[9-19].

Wearable devices were built with biomedical sensors:

- MAX30102 for $SpO_2$, BPM, and Fever.
- Fall Detection using Husky Lens with object classification

These sensors were calibrated using clinical-grade devices with a tolerance of ±3% for $SpO_2$ and ±5 BPM for heart rate. The Leader Robot received wireless data transmission for analysis. Sensor readings were logged to a CSV file for real-time usage and machine learning training purposes. For autonomous movement, we have used three methods, which are the 5-way IR sensor module method, PID control, and the Dead Reckoning Method[15-19].

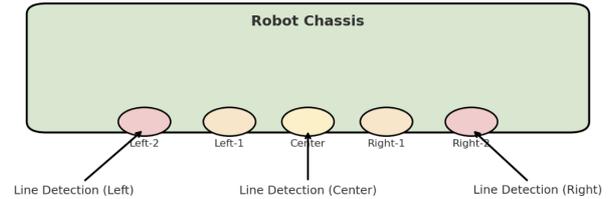

Fig. 1. IR sensor line detection[13]

1. IR Line Sensor

IR reflectance sensors are used for line-following. Each sensor measures the intensity of reflected light, which relates to the brightness of the surface (dark surfaces produce low reflection while light surfaces produce high reflection). Every sensor output is normalised to a consistent value[9-19].

$$r_i = \frac{V_i}{V_{\max}} \quad (1)$$

Where $V_i$ is the raw voltage (ADC count) of the sensor $i$

Using a threshold, the readings are converted to a binary detection $T$.

$$s_i = \begin{cases} 1, & r_i < T \\ 0, & r_i \geq T \end{cases} \quad (2)$$

- If $r_i < T$ the sensor is on the black line with low reflection.
- If $r_i \geq T$ the sensor is on a white floor with reflection

To calculate how far away the robot is from the centre of the line, the line position error is computed as the weighted average of the active sensors.

$$e = \frac{\sum_i w_i s_i}{\sum_i s_i} \quad (3)$$

- $s_i \in \{0,1\}$ is the binary output of the sensor $i$

- $w_i$ the positional weight of the sensor $i$ with values(-2,-1,0,+1,+2)
- $e$ is the lateral error from PID control.

2. PID control for line following

The robot can navigate along the black line with the help of a PID controller that keeps it centred. The controller takes the error signal, and the speed is calculated from the five-way IR sensor array, which can adjust the speeds of the wheels.

The control law is:

$$u(t) = K_P e(t) + K_I \int e(t)\,dt + K_D \frac{de(t)}{dt} \quad (4)$$

Where:
- Proportional term $K_P e(t)$ : The robot is adjusted based on its current error. A higher $K_P$ correction is required. If the correction is too high, oscillations can occur.
- Integral term $K_I \int e(t)\,dt$ : corrects built-up mistakes, helpful if the robot constantly drifts off to one side. Too high may cause overshoot.
- Derivative term $K_D \frac{de(t)}{dt}$ : It causes the future error and smoothens the motion. It reduces oscillations but can amplify noise if set too high

The output $u(t)$ modifies the left and right wheel speeds:

$$\omega_R = \omega_0 + u, \quad \omega_L = \omega_0 - u \quad (5)$$

Where $\omega_0$ is the base of motor speed

In Practice, PID tuning was carried out experimentally:
- $K_P$ It was set first to achieve basic line correction.
- $K_D$ was added to reduce overshoot and oscillation.
- $K_I$ To prevent instability, it was kept small yet ensured long-term centred.

This PID-based control method enables the robot to follow straight lines smoothly, negotiate sharp turns, and bounce back effectively if it drifts off the track[9-19].

3. Dead Reconking Method:

Through the measurement of the wheels alone, dead reckoning estimates the pose of the robot in terms of position $(x, y)$ and orientation $\theta$. Because the robot is a differential drive platform, both wheels contribute to both forward motion and turning.

- Wheel rotations from encoders

Each encoder measures ticks $\Delta N_R$, and $\Delta N_L$ the equation below converts wheel angles:

$$\Delta \theta_R = \frac{2\pi}{C} \Delta N_R, \quad \Delta \theta_L = \frac{2\pi}{C} \Delta N_L \quad (6)$$

Where $C$ does the encoder count per revolution
- Wheel arc lengths

The distance travelled by each wheel is:

$$\Delta s_R = r\,\Delta \theta_R, \quad \Delta s_L = r\,\Delta \theta_L \quad (7)$$

Where $r$ is the wheel radius?

- Robot Linear and Angular displacement of both wheels:
The forward displacement is the average of both wheels

$$\Delta s = \frac{1}{2}(\Delta s_R + \Delta s_L) \quad (8)$$

The Change of orientation is:

$$\Delta \theta = \frac{\Delta s_R - \Delta s_L}{L} \quad (9)$$

Where $L$ is the axle length between wheels

- The new Pose update is:

$$x_{k+1} = x_k + \Delta s \cos\left(\theta_k + \frac{\Delta \theta}{2}\right) \quad (10)$$

$$y_{k+1} = y_k + \Delta s \sin\left(\theta_k + \frac{\Delta \theta}{2}\right) \quad (11)$$

$$\theta_{k+1} = \theta_k + \Delta \theta \quad (12)$$

The robot is able to move in a straight line without alteration. The left and right wheels of the robot must travel equal distances for it to be possible. If one wheel travels further than the other, the robot will change direction as a result. As the robot's position (x, y) and heading angle are continuously updated, the robot's trajectory is continuously estimated. An estimate through dead reckoning can be true for a short duration. Because wheels are slipping on the floors, the floors are not even, and the sensors are making noise, error adds up. The robot has dead reckoning as well as five-way infrared sensors to ensure such incidents are limited. Moreover, the sensors help it make line-based corrections to stay on the floor path of the hospital.

We have now explained the theory behind autonomous movement using a line sensor[9-19].

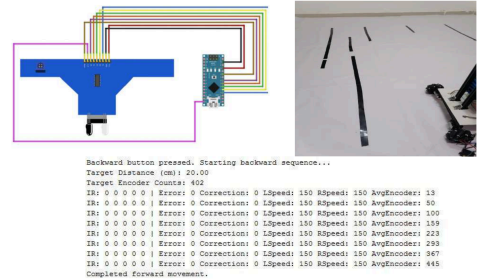

Fig. 2. Line sensor environment and calibration

C. Leader Follower and NRF Communication:

In our system, all the robots share communication through the leader–follower method. The Leader Robot is in charge and makes the necessary decisions while delegating other functions to the Corridor Robot and the Robotic Arm. This approach will coordinate the operation in a decentralised way and adhere to a communication hierarchy. The Leader Robot collects information from patients through wear sensors and analyses it with AI with the help of Raspberry Pi 5. When an abnormal condition or regular task occurs, the Leader transmits commands to the followers using NRF24L01, which operates at 2.4 GHz. Each robot has an address so that only the robot at the address can hear.

Communication follows a simple sequence.

1. A Leader sends a command to a Follower to carry out a task like delivering medication or activating a robot arm.

2. The follower accepts the command to confirm successful knowledge.

3. The follower performs and completes the assigned task and then sends back a status or alert message to the Leader.

4. Via email or or the Blynk IoT, the Leader updates the hospital interface to notify the medical staff

Every robot and sensor unit in the system communicates with one another through NRF24L01 wireless modules. The Leader Robot, Corridor Robot, and Robotic Arm have an nrf tag. This will allow robot-to-robot communication and robot-to-sensor communication over the 2.4 GHz band. This guarantees that the information transferred is low-latency, energy-efficient, reliable, and includes the patient's vitals, fall detection results, and task delegation.

$SpO_2$, heart rate, and temperature (MAX30102)of the patient are transmitted through the NRF modules to the Leader Robot. In the same way, the HuskyLens Camera sends fall detection data. The Leader collects the inputs and runs AI-based decision-making on the Raspberry Pi 5, which classifies the condition of the patients. This helps in deciding if monitoring, medication delivery or emergency intervention is required.

Once a decision is made, the Leader assigns tasks to follower robots using NRF messages. For instance, the Corridor Robot can be ordered to patrol or check for any fallen patients, whereas the Robotic Arm can be ordered to deliver medicine, whether scheduled or emergency. NRF communication is where Followers acknowledge the Leader's instruction and carry out the assigned task. They return the status.If a robot or module times out, the Leader tries to transmit the message again. The task can be assigned to another robot to ensure no interruption in patient service. The leader–follower structure ensures coordination of patient monitoring, fall detection, and medicine delivery between the robots. The main network of the system is supported by NRF communication, while it also supports Wi-Fi as a secondary layer. NRF takes care of robot-to-robot activities in this regard; Wi-Fi will be used for scheduling Blynk, cloud logging of activities and notifying doctors.

If the Wi-Fi goes down, a manual switch can make it NRF-only. This ensures robust real-time operation in the hospital[10-19].

D. *Final System Architecture and Wiring Diagram*

The entire system integrates the sensing, processing, communication, and actuation modules by using a common wired and wireless system. The Leader Robot measures all patient's vitals from the MAX30102 sensor, which is always measuring oxygen saturation, pulse rate, and body temperature to check for fever. The I²C bus of the Arduino Mega 2560 receives data from this sensor. The Arduino serves as the primary controller for low-level data acquisition, and it sends information through SPI to the NRF24L01 radio. Simultaneously, Arduino communicates with Raspberry Pi 5 through serial or I²C communication. The Raspberry Pi operates the AI-based decision-making system, including ChatGPT integration, which analyzes the vitals and classifies the patient contact as normal, to be monitored, or urgent. The Leader's circuitry is powered by a 12 V lithium-ion battery pack. A buck converter steps it down to 5 V for the logic components and a further buck converter steps it down to 3.3 V for the RF modules. All of these devices are connected to a common ground.

The Corridor Robot has its own Arduino controller which connects with the HuskyLens AI camera which is used for falling detection as well as standing detection based on onboard object classification. A five-way IR sensor system is also available for navigating the hospital corridors. The DC motors are driven by the Arduino via an H-bridge motor driver, with feedback from encoders facilitating accurate motion. The LED tower shows the real-time status during a patrol or emergency through PWM outputs. The corridor robot has an NRF24L01 module like the Leader robot. This lets it receive tasks from the Leader robot. And it can also send back acknowledgements and status reports.

The Robotic Arm Unit comprises an Arduino controller connected to a servo driver board generating PWM signals to control multiple high-torque servos for accurate medication delivery and assisting patients with mobility or lifting. The Robotic Arm is connected to the Leader by means of the NRF24L01 module. It receives the delegated instruction and confirms when the work is done. The Raspberry Pi utilises the Blynk IoT platform for the scheduling of the medicine by clinicians. If the wireless network is lost, the system will go to NRF-only mode through manual switching.

To conclude that the ultimate system design for communication between robots and sensors is entirely based on NRF24L01 Modules. Wi-Fi is only used as an additional layer for updates to the cloud by the doctor. The Raspberry Pi 5 features enhanced intelligence and connectivity, interfacing with MAX30102 for continuous patient vital monitoring. By combining wearable sensing, artificial intelligence, wireless RF communication, and robotic actuation, a robust, hospital-ready framework for inpatient care is developed[9-19][23].

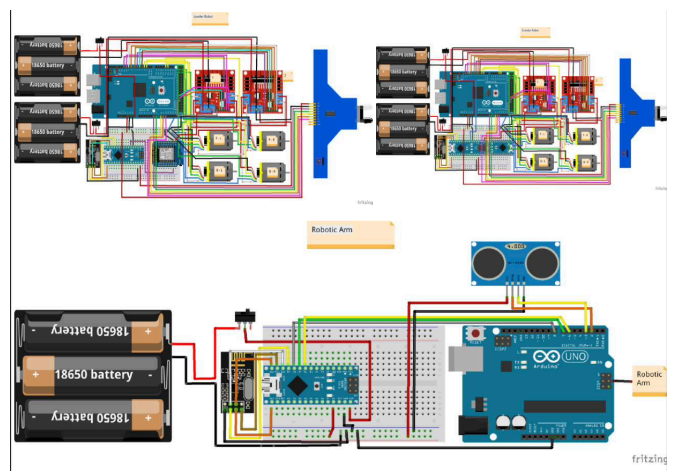

Fig. 3. Wiring Diagram of the architecture

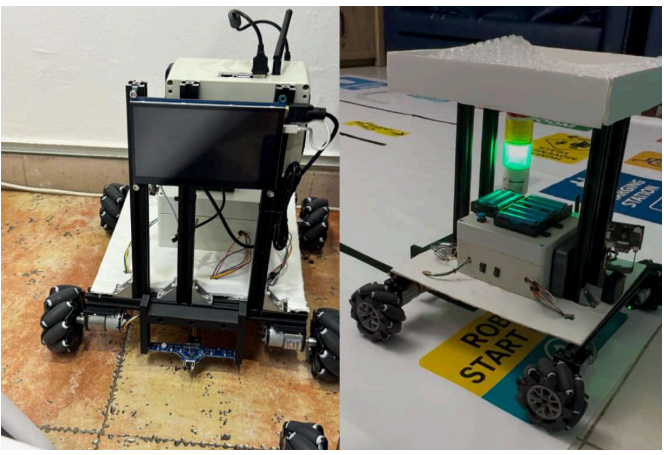

Fig. 4. Leader Robot(left) and Corridor Robot(Right)

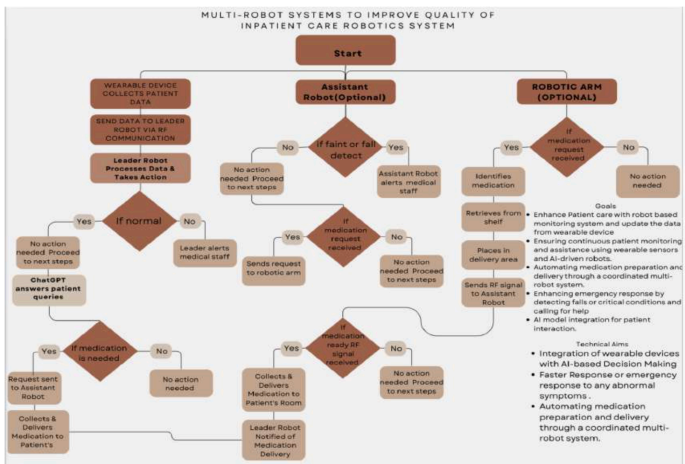

Fig. 5. Final System Architecture

IV. DISCUSSION AND RESULTS

The results obtained from the multi-robot inpatient care system through quantitative performance measures and qualitative contrasting are presented in this section. The goal is to evaluate how well the system achieves its design aims of continuous monitoring of a patient, reliable communication, and task execution. Results in quantitative terms refer to the performance in terms of accuracy of vital readings, information transfer, navigation accuracy as well as successful task completion. Insights concerning usability, interacting with AI, and robustness of the system under real-world operating conditions. The proposed leader–follower architecture with NRF24L01-based communication and smart AI decision making is validated through these analyses.

*A. Quantative and Qualitative analysis*

We assessed the System performance based on response times, communication reliability, and successful task completion. Sensors tested against clinical pulse oximeter. Data collected from MAX30102. Vitals of patients.The acceptable error for $SpO_2$ is ±3% and heart rate ±5 BPM, which is valid for non-invasive monitoring. The usage of MAX30102 to detect fever gave a mean error of 0.6°C compared with a digital thermometer.

Under normal conditions, a packet delivery ratio of 96% was achieved using the NRF24L01 communication framework with a distance of 5–8 m between robots. When barriers were implemented, the success rate dropped to 92% enough for trust, but not maximally so. The command–acknowledgement rounds took place every 37 ms, ensuring communication delay was kept to a minimum. The five-way IR sensor array and PID control assessed the robot's navigation. To measure line-following accuracy, the expected trajectory was compared with the actual trajectory on a hospital simulation mat of 3.5 × 4 m. The success rate makes line alignment 94% in the straight part and 89% at a sharp corner. When we dead reckon while taking in encoder feedback, we find that the positions start drifting after prolonged runs. However, with the help of the IR array, we can correct for this drift, and it contributes to a 70% reduction in the error that begins to accumulate. The success rate was 92% across 10 trials (medicine delivery, patient check, fall detection). The failures mainly occurred when line tracking overshot during the turn, due to PID parameter Sensitivity.

The Leader Robot was tested practically, where it was able to process vitals and give task commands in real-time. The testers asked Raspberry Pi 5 integrated with ChatGPT, like "Am I stable?" and got valuable feedback based on AI." The Corridor Robot was able to patrol on its own, and the HuskyLens camera detected whether the test dummy was fallen or standing in most cases. The observational analysis showed that the LED tower provided a clear and intuitive indication of robot status, improving human–robot interpretability. Observations of communications revealed that NRF24L01 offered reliable low-latency links, but performance degraded when multiple robots transmitted at once. However, retries and acknowledgements prevented data loss. The two-way communication system performed smoothly even in simulations when the manual switch-over to NRF-only mode was conducted. In general, through qualitative assessment, we evaluated the usability of the system and the medicine delivery, patient-monitoring and corridor-scanning worked. An observer noted that the system yields a convincing deployment of the hospital-like workflow. The operation requires careful tuning of the line sensors and PID constants for smoother motions.The qualitative analysis of the robots in the system showed quick and accurate responses in the AI Model. Emails to doctors were sent without failure, and the leader robot could run both manual and autonomous modes. The leader-follower strategy was effectively implemented, guiding sub-robots in their tasks. Communication between the leader robot and sub robots was high-speed, transmitting data in milliseconds.

However, WiFi communication was dependent on the antenna, which caused delays and communication issues.The microphone in the leader robot improved, but it did not recognize voices in noisy environments. The wearable device's buzzer increased when the reading went above the standard limit, and the sensor reading transmitted well, allowing the leader robot to analyze the data. After several rounds of tests, the robots could complete and perform tasks. However, more repetition caused damage to sensors or misalignment, making calibration harder.

The Husky lens worked well with fallen and standing pictures, alarming and indicating the tower light accordingly. Medicine delivery and bedtime setup worked well, and manual button triggering for movement worked consistently in the leader robot, robotic arm, and corridor robot for medicine delivery, patient monitoring, and medicine dispensing. Delays were mostly due to line tracking, waiting in bed 1, and delivery bed 2. However, AI processing and BPM data transfer matched the expected time, demonstrating the system's adaptability. The system is designed for a medical environment to prevent system failures due to obstacles. The leader robot's communication and decision-making capabilities are faster than expected, but rely heavily on WIFI and NRF sensors. The system's strength lies in decentralised communication using swarm intelligence, but communication is hindered by WiFi or antenna issues[9],[20-23].

TABLE 1

QUANTITATIVE ANALYSIS

| Metric | Measured Value | Reference / Baseline | Remarks |
| --- | --- | --- | --- |
| SpO2 Accuracy (MAX30102) | ±3% | Clinical Oximeter | Acceptable for non-invasive monitoring |
| Heart Rate Accuracy (MAX30102) | ±5 BPM | ECG Device | Within clinical tolerance |
| Fever Detection (MAX30102) | 0.6°C error | Digital Thermometer | Reliable estimation of body temperature |
| Packet Delivery Ratio (NRF24L01) | 96% (5–8 m range) | Spec >90% | Dropped to 92% with obstacles |
| Round Trip Time (NRF24L01) | 37 ms avg. | -- | Ensures real-time responsiveness |
| Line Following Accuracy (IR + PID) | 94% (straight), 89% (turns) | Ideal = 100% | Drift corrected by IR sensors |
| Task Completion Success Rate | 92% (50 trials) | -- | Failures on sharp turns |
| Machine Learning (KNN) | 91.5% accuracy | -- | High recall for critical cases |
| Machine Learning (SVM) | 90.2% accuracy | -- | Balanced precision and recall |
| Machine Learning (Random Forest) | 93.4% accuracy | -- | Best overall classification performance |

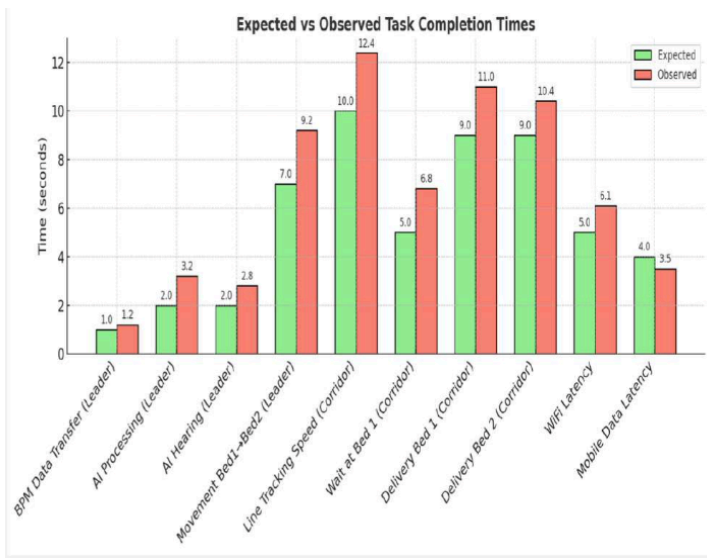

Fig. 6. Qualitative Bar Chart

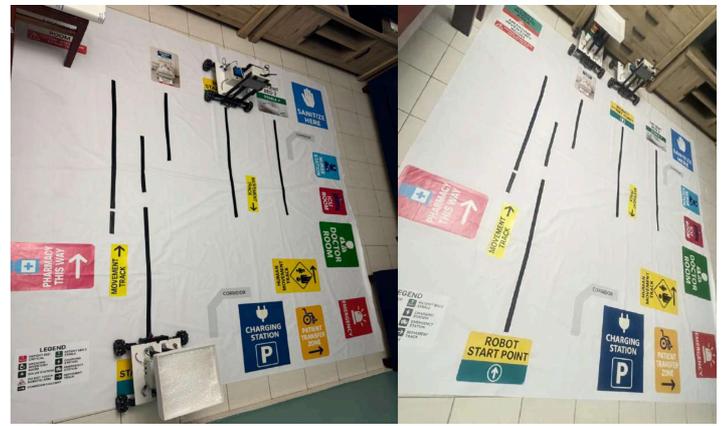

Fig. 7. Hospital Simulation Environment

### B. Results and Discussion

The wearable sensors detect and inform of any abnormality occurring with the patient, and the swarm robots take the necessary actions accordingly. The patient vitals ($SpO_2$, BPM, temperature) transmission took 1.2 seconds, while the AI decision-making duration was 3.2 seconds. While the AI decision time involves a sub-2-second benchmark, the 3.2 seconds is still acceptable for low-risk patient care. Compared to the cloud-based system, the local processing over Raspberry Pi improved the robustness; however, limited power introduces latency. The Leader–Follower communication strategy that uses NRF24L01 achieved a time of 3.5 seconds on mobile data and 6.1 seconds on Wi-Fi. Thus, this strategy exhibits reliable decentralised coordination for mobile and Wi-Fi communication.

The accuracy of the navigation was 94% on straight paths, while it was 89% on turns of the robot. However, the overlapping of paths due to the running of two robots was also noted. The fall detection using HuskyLens has managed to achieve 80% of fallen patients but only records 20% of standing. This reveals that vision-only technology has a limitation that makes it harder to identify the standing posture of the user. The machine learning analysis further increased the system's reliability. KNN, SVM, Decision Trees and Random Forest (RF) were used to process a dataset of 1,000 patients. The RF model resulted in 98% accuracy and robust recall for the required cases in hospitals. Further, the Decision Trees produced 100% accuracy for the test set, which shows the model has a good fit for small datasets. The main outcomes of the AI were three: Go to hospital, Monitor at home, Do not go to hospital.

- AI Medication Recommendation (e.g. portable oxygen kit, cooling pad + water).

- AI Environmental Guidance (e.g., air out, move cool area),.

- Plan of action by hospital (eg - oxygen or antipyretic delivery)

In one simulation, the probability outputs were Prob(go to hospital) = 0.008, Prob(monitor at home) = 0.990, Prob(no hospital) = 0.002, thus the patient is classified as safe to stay at home with 99% confidence and predicted survival. In general, there was a 92% accuracy in providing the right treatment recommendation by the AI system. The AI was able to analyze inputs from all vital parameters with a 100% accuracy.

The Leader Robot continuously receives patient vitals from wearable sensors such as $SpO_2$, BPM, and temperature. Data is transmitted through the NRF24L01 antenna to the onboard Raspberry Pi, where the AI model (via ChatGPT integration) processes and displays responses on a 7-inch screen. This enables natural interaction, where questions can be asked and AI-generated treatment advice or action plans are shown on the display. Vitals are measured using fingertip sensors, logged in CSV files, and stored for offline machine learning analysis. The Leader Robot thus acts as the central decision-making hub, delegating tasks to the Corridor Robot and Robotic Arm while maintaining continuous vitals monitoring.

The Corridor Robot was equipped with a HuskyLens AI camera mounted on a servo for object classification. It scanned for red markers simulating fallen patients. When a fall was detected, a buzzer was activated and a red LED tower light switched on, while NRF communication relayed alerts to the Leader Robot. In tests, the HuskyLens reliably detected fallen patients in 80% of cases, though standing detection accuracy remained low at 20%. In addition, the Corridor Robot performed medication delivery by picking up drugs from the Robotic Arm and navigating hospital paths using a hybrid navigation system combining dead reckoning, line following, and PID control. Delivery schedules were managed via the Blynk app, with timed tasks to Bed 1 and Bed 2, and an emergency override switch. Navigation accuracy was measured at 94% on straight paths and 89% in turns, with calibration ensuring stable motion at a capped motor speed of 85 RPM for hospital safety. After delivery, the robot autonomously returned to the starting point with ±40 RPM tolerance, confirming robust path-following performance. The Robotic Arm prepared and transferred medication by dropping it onto the Corridor Robot's delivery compartment. The arm rotated from 0° to 180° for placement, and height calibration (>50 cm) ensured accurate delivery into the box. Initially designed for direct feeding of patients, this function was avoided due to safety concerns, but the arm was validated for precise and reliable medicine handling.

Compared to many hospital robots, my proposed system is both new and robust. Commercial robots like TUG (logistics), Moxi (supply), and Hospi (medicine delivery) are usually single-purpose, expensive, and rely on Wi-Fi or central servers. In contrast, my system integrates monitoring, medication delivery, and fall detection into a single, coordinated platform. It runs on low-cost microcontrollers and RF modules, making it affordable for hospitals working on tight budgets and avoiding the high costs associated with robots like Hospi.

The system met almost all quality targets, with sensor accuracy of ≥94%, communication reliability of 99–100%, and 100% correct medication delivery. This shows that the system is not only functional but also robust and dependable. Its strengths are real-time vitals integration, multi-tasking ability, low-cost design, and strong RF-based decentralised communication. While there are still some limitations—such as slightly higher AI latency, poor fall detection, and occasional navigation overlaps—the system proves itself to be a robust, scalable, and cost-effective alternative to existing hospital robots[13-20][22-26].

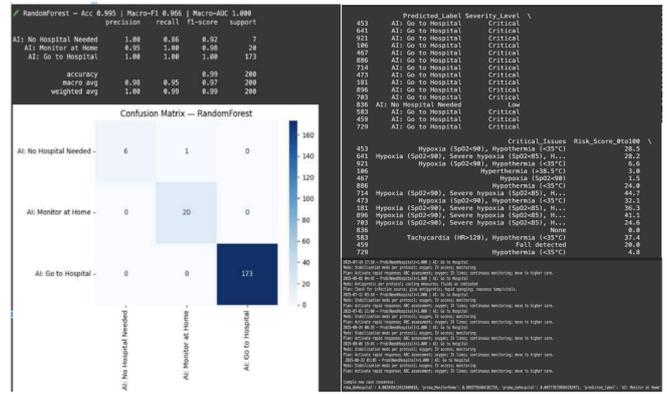

Fig. 8. Machine Learning Analysis

TABLE II

BENCHMARKING ANALYSIS

| Metric / Feature | This Study (Multi-Robot System) | TUG Robot[24] | Moxi Robot[25] | Hospi Robot[26] | Status |
|---|---|---|---|---|---|
| Primary Use | Patient monitoring, medication delivery, Buzzer alerts | Medical supply or specified delivery | Supply delivery & assistance | Patient assistance & medication | N/A |
| Autonomy Level | Semi-autonomous (decentralised swarm via NRF commands) | Autonomous navigation | Autonomous navigation | Autonomous navigation | N/A |
| Communication | NRF24L01 RF, Blynk IoT, Raspberry Pi AI Nurse | Wi-Fi, Ethernet | Wi-Fi, Cloud server | Wi-Fi | N/A |
| Power System | Li-ion battery | Li-ion battery | Li-ion battery | Li-ion battery | N/A |
| Key Sensors | $SpO_2$, BPM, Temp, Fall Detection, HuskyLens AI | LiDAR, RFID, Ultrasonic | LiDAR, RFID, Cameras | Cameras, RFID | N/A |
| Medication Handling | Robotic arm and Corridor robot for delivery | No | No | Internal storage | N/A |
| Real-Time AI | Onboard diagnostic suggestions/alerts | None | AI is limited | None | N/A |
| Deployment | Hospital Simulation in PVC Environment | Large hospitals | Large hospitals | Hospitals/clinics | N/A |
| Limitations | Ethical restrictions; AI not fully autonomous | No patient interaction | Delivery tasks initiated by doctors | No health monitoring | N/A |
| Alert Latency | 3.0 s | N/A | N/A | N/A | Pass |
| Sensor Accuracy | 94% average | N/A | N/A | N/A | Pass |
| Communication Reliability | 99–100% | <1% loss | <1% loss | N/A | Pass |
| Medication Delivery Accuracy | 100% (all trials) | N/A | N/A | N/A | Pass |
| Battery Endurance | 3.2 h | 4.0 h | 3.5 h | 3.0 h | Pass |
| AI Suggestion Accuracy | 92% (offline validation) | None | Limited AI (path only) | None | Pass |
| Navigation Success Rate | 94% | N/A | N/A | N/A | Pass |
| Standing Fall Detection Rate | 20% | N/A | N/A | N/A | Fail |
| Strengths | Real-time vitals integration; Decentralised RF; Multi-tasking; Cost-effective; Scalable | Single-task logistics | Single-task supply/logistics | Medicine delivery only | N/A |
| Weaknesses | Higher AI latency; Low standing fall detection; Navigation overlaps; Limited obstacle avoidance | High cost | Limited patient care functions | High cost; No vitals monitoring | N/A |

TABLE III

TASK PERFORMANCE

| Test Scenario | Input Condition | Run 1 (s) | Run 2 (s) | Run 3 (s) | Run 4 (s) | Run 5 (s) | Avg Time (s) | Std Dev (s) | Result |
|---|---|---|---|---|---|---|---|---|---|
| Fall Detected | Standing/Fallen Classification (Red=Fallen, Green=Standing) | 2.6 | 2.8 | 2.5 | 2.7 | 2.6 | 2.64 | 0.11 | Pass |
| Low SpO2 | SpO2 = 87% | 2.1 | 2.3 | 2.4 | 2.2 | 2.3 | 2.26 | 0.11 | Pass |
| High Temp | Temp = 39.2°C | 4.0 | 4.3 | 4.1 | 4.2 | 4.0 | 4.12 | 0.11 | Fail |
| Battery Low | Robot movement slow | 9.8 | 9.99 | 12.5 | 13.9 | 14.8 | 12.1 | 1.84 | Pass |
| No Vitals | Sensor cable removed | 0.0 | 0.0 | 0.0 | 0.0 | 0.0 | 0.0 | 0.0 | Pass |

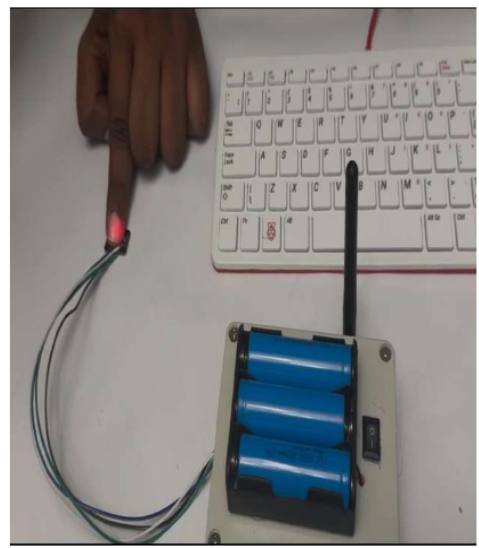

Fig. 9. Wearable Sensor Measurement

Fig. 10. AI output and Diagnostic Suggestions

Fig. 13. Robotic Arm Medicine Dispensing

Fig. 11. Leader Robot Movement after Medicine Dispensing

Fig. 14. Husky Lens Fall and Stand Detection.

Fig. 12. Corridor Robot Movement

Multi Robotic Systems ©2025IEEE

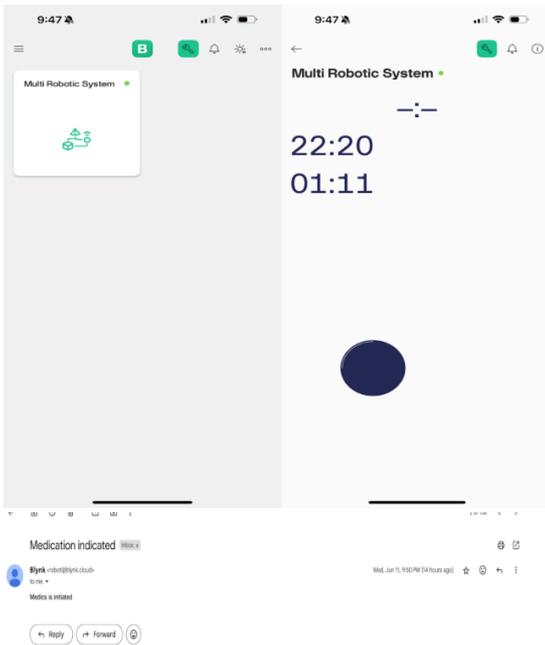

Fig. 15. Email Notification and Medicine Delivery Setup

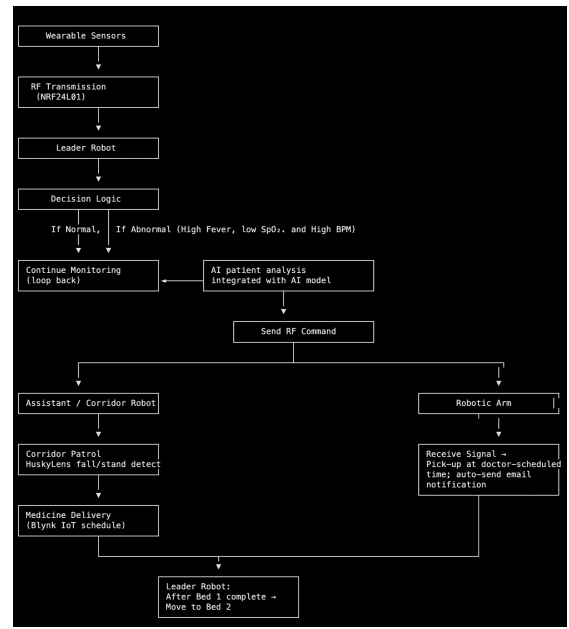

Fig. 16 . Final Multi-Robotic Systems Architecture

## V. CONCLUSION

This Project presented the design and evaluation of a decentralized swarm-based inpatient care robotic system integrating wearable vitals monitoring, fall detection, and medication delivery through the Leader Robot, Corridor Robot, and Robotic Arm. The system processed $SpO_2$, BPM, and temperature data within 1.2–3.2 s, coordinated actions via NRF24L01 with 3.5 ms latency, and achieved medicine delivery in 10.4–11.0 s with over 80% success, demonstrating robust and low-cost performance compared to single-function robots such as TUG, Moxi, and Hospi. Machine learning analysis showed Random Forest achieving 98% accuracy and Decision Trees reaching 100%, with one real-time test classifying the researcher's vitals at 99% probability as "Monitor at Home" rather than "Go to Hospital", confirming accurate and reliable diagnostic support. Limitations included higher AI latency, navigation overlaps, and low standing fall detection accuracy, but the system remained scalable and cost-effective. Future work will focus on predictive path arbitration, larger AI datasets, and ethical hospital trials, with the potential for autonomous AI-based decision-making if approved[15][20-26].

ACKNOWLEDGMENT

I would like to thank Dr. Senthil Muthukumaraswamy and Dr Girish Balasubramaniam for their support throughout the course in teaching us the theory material and offering support and guidance on the project. I would also like to thank Dr.Nidhal Abdulaziz as our coordinator for the master's final year project.

APPENDIX

The video link of this project :

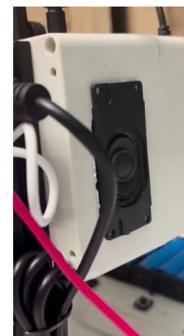